\documentclass[11pt,twoside]{article}
\usepackage[utf8]{inputenc}
\usepackage{graphicx}
\usepackage{amsmath, amssymb, bm}
\usepackage{hyperref}
\usepackage{natbib}
\usepackage{fancyhdr}
\usepackage{geometry}
\usepackage{titlesec}
\usepackage{lipsum}
\usepackage{orcidlink}
\usepackage{booktabs}
\usepackage{placeins}
\usepackage{float}
\usepackage{placeins}
\setcitestyle{authoryear}
\usepackage{amsmath,amssymb,amsfonts,bm,mathtools}
\usepackage[ruled,vlined,linesnumbered]{algorithm2e}

\newcommand{\R}{\mathbb{R}}
\newcommand{\E}{\mathbb{E}}

\hypersetup{
colorlinks=true,
linkcolor=blue,
filecolor=blue,
citecolor=blue,
urlcolor=black
}
\fancypagestyle{plain}{
\fancyhf{}

}
\title{
\vspace{-1.5em}
\hrule height 1.5pt
\vspace{0.8em}
AR-Flow VAE: A Structured Autoregressive Flow Prior Variational Autoencoder for Unsupervised Blind Source Separation
\vspace{0.8em}
\hrule height 1.5pt
\vspace{1em}
}

\author{%
\begin{minipage}[t]{.48\textwidth}\centering\small
  \textbf{Yuan-Hao Wei\orcidlink{0000-0001-9439-0780}}\\
  Hong Kong Polytechnic University \\
  \texttt{Yuan-Hao.Wei@outlook.com}
\end{minipage}\hfill
\begin{minipage}[t]{.48\textwidth}\centering\small
  \textbf{Fu-Hao Deng\orcidlink{0000-0001-8034-6949}}\\
  Wuhan University of Technology \\
  \texttt{dengfuhao@whut.edu.cn}
\end{minipage}\\[3em] 
\begin{minipage}[t]{.48\textwidth}\centering\small
  \textbf{Lin-Yong Cui\orcidlink{0009-0005-9142-1728}}\\
  The University of Hong Kong \\
  \texttt{cuilinyo@hku.hk}
\end{minipage}\hfill
\begin{minipage}[t]{.48\textwidth}\centering\small
  \textbf{Yan-Jie Sun\orcidlink{0000-0002-7967-6382}}\\
  Hong Kong Polytechnic University \\
  \texttt{Yanjie.Sun@connect.polyu.hk}
\end{minipage}
}

\date{}
\begin{document}
\maketitle
    \thispagestyle{plain}
    \begin{abstract}
        Blind source separation (BSS) seeks to recover latent source signals from observed mixtures. Variational autoencoders (VAEs) offer a natural perspective for this problem: the latent variables can be interpreted as source components, the encoder can be viewed as a demixing mapping from observations to sources, and the decoder can be regarded as a remixing process from inferred sources back to observations. In this work, we propose AR-Flow VAE, a novel VAE-based framework for BSS in which each latent source is endowed with a parameter-adaptive autoregressive flow prior. This prior significantly enhances the flexibility of latent source modeling, enabling the framework to capture complex non-Gaussian behaviors and structured dependencies, such as temporal correlations, that are difficult to represent with conventional priors. In addition, the structured prior design assigns distinct priors to different latent dimensions, thereby encouraging the latent components to separate into different source signals under heterogeneous prior constraints. Experimental results validate the effectiveness of the proposed architecture for blind source separation. More importantly, this work provides a foundation for future investigations into the identifiability and interpretability of AR-Flow VAE.
    \end{abstract}

    
\section{Introduction}

    Blind source separation (BSS) aims to recover latent source signals from observed mixtures without prior knowledge of the mixing process, and independent component analysis (ICA) has long provided one of its most influential theoretical and algorithmic foundations (\cite{Comon1994,BellSejnowski1995,Hyvarinen1999FastICA,HyvarinenOja2000}). In the classical setting, identifiability is closely tied to statistical independence and non-Gaussianity of the latent sources \cite{Comon1994,HyvarinenOja2000}. However, once the problem moves beyond linear mixtures and simple source models, the separation task becomes substantially more challenging; in particular, unrestricted nonlinear ICA is generally non-identifiable, which limits the direct applicability of naive nonlinear latent-variable models to BSS (\cite{HyvarinenPajunen1999}). 
    
    A recent development is that nonlinear ICA and nonlinear BSS can become identifiable again when additional structure is introduced, such as temporal dependencies, nonstationarity, or auxiliary observed variables (\cite{HyvarinenMorioka2016,HyvarinenMorioka2017,HyvarinenSasakiTurner2019,Khemakhem2020IVAE,Hyvarinen2023Patterns}). These advances have also reshaped the interpretation of variational autoencoders (VAEs) in source recovery problems: rather than treating latent variables as abstract compressed features, one may regard them as source components whose separation is promoted by suitable structural assumptions in the latent space (\cite{Khemakhem2020IVAE,Hyvarinen2023Patterns}). This line of work suggests that the success of deep latent-variable models for BSS depends not only on expressive encoders and decoders, but also on whether the latent prior encodes the right form of source structure (\cite{KingmaWelling2014}). 
    
    Motivated by this perspective, VAE-based BSS has recently attracted increasing attention (\cite{Neri2021VAEBSS,DeSalvio2023VAEAcoustics}). In such models, the encoder can be interpreted as a demixing map from observations to latent sources, while the decoder plays the role of a remixing or generative mixing map back to the observation space (\cite{KingmaWelling2014,Neri2021VAEBSS}). At the same time, the broader VAE literature has shown that replacing the standard factorized Gaussian prior with structured priors---such as graphical-model priors, state-space priors, or Gaussian-process priors---can substantially improve latent modeling when dependencies across time, samples, or latent variables are important (\cite{Johnson2016SVAE,Karl2017DVBF,Hafner2019PlaNet,Casale2018GPPriorVAE,Fortuin2020GPVAE,Jazbec2021ScalableGPVAE,Connor2021LatentStructure,ZhaoLinderman2023SVAE,Zhu2023MarkovGPVAE,wei2024innovative,wei2025structured}). In parallel, normalizing flows and autoregressive flows have emerged as powerful tools for constructing flexible latent distributions with exact change-of-variables likelihoods (\cite{RezendeMohamed2015NF,Kingma2016IAF,Papamakarios2017MAF,Huang2018NAF,Dinh2017RealNVP,KingmaDhariwal2018Glow,Papamakarios2021Flows}). 
    
    Nevertheless, existing nearby strands still leave room for a more direct combination of these ideas in BSS. Some prior VAE-based BSS studies have relied on standard latent priors or comparatively rigid structured priors (\cite{Neri2021VAEBSS}), while much of the normalizing-flow literature in VAEs has focused on posterior approximation, generic density modeling, or conditional structured prediction rather than on per-source structured priors for blind source separation (\cite{RezendeMohamed2015NF,Kingma2016IAF,Papamakarios2017MAF,Huang2018NAF,Bhattacharyya2019CFVAE}). To the best of our knowledge, the explicit use of a \emph{per-source autoregressive flow prior} inside a VAE for BSS has not been systematically established in prior work. This motivates the present study, in which we propose \textbf{AR-Flow VAE}, a structured VAE framework where each latent dimension is treated as a source component and assigned a parameter-adaptive autoregressive flow prior. The resulting model preserves exact prior-density evaluation while substantially enriching the latent source distribution, allowing it to capture complex non-Gaussian behaviors and structured dependencies, such as temporal correlations, beyond those representable by conventional Gaussian or linear autoregressive priors. More importantly, this work is intended not only to validate the BSS capability of the proposed architecture, but also to provide a basis for subsequent studies on the identifiability and interpretability of AR-Flow VAE itself (\cite{wei2026nonlinear}).

\section{Structured Priors in VAEs}

A broad body of work has shown that replacing the standard isotropic Gaussian prior in a VAE with a structured latent model can significantly improve representation quality when dependencies matter (\cite{Johnson2016SVAE,Hafner2019PlaNet,Casale2018GPPriorVAE,Fortuin2020GPVAE,Jazbec2021ScalableGPVAE,Connor2021LatentStructure,ZhaoLinderman2023SVAE,Zhu2023MarkovGPVAE}). Structured VAE formulations combine graphical-model priors with neural observation models (\cite{Johnson2016SVAE,ZhaoLinderman2023SVAE}), while sequential and dynamical variants such as DVBF and recurrent state-space models bring latent transition structure into deep generative learning \cite{Hafner2019PlaNet}. Gaussian-process-based priors further enrich VAE latent spaces by modeling correlations across time or samples (\cite{Casale2018GPPriorVAE,Fortuin2020GPVAE,Jazbec2021ScalableGPVAE,Zhu2023MarkovGPVAE}), and related work has explored learned latent structures beyond standard GP formulations (\cite{Connor2021LatentStructure}). These developments strongly support the general principle that better structured priors can improve latent-variable modeling, but they mostly rely on Gaussian, state-space, or message-passing-based structures rather than autoregressive normalizing-flow priors. 

\begin{figure}[t]
    \centering
    \includegraphics[width=0.95\linewidth]{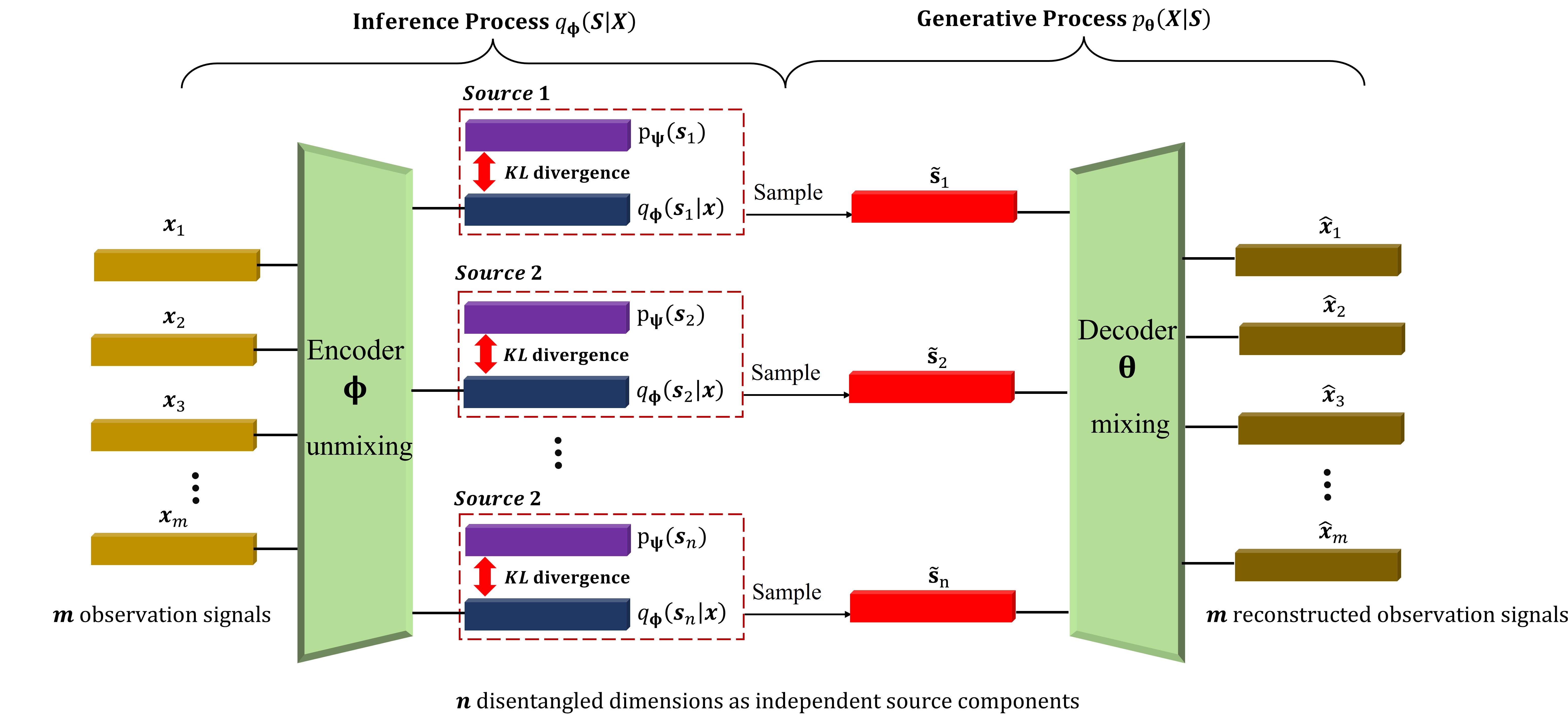}
    \caption{Typical structured-prior VAE for BSS. The encoder is interpreted as a demixing operator from observed mixtures to latent source components, while the decoder acts as a generative mixing operator that reconstructs the observations from the inferred sources. Each latent dimension is assigned an individual parameterized prior, and the corresponding posterior is regularized toward that prior through the KL divergence.}
    \label{fig:structured_prior_vae_bss}
\end{figure}

A closely related line of our previous studies has explored \emph{structured-prior VAEs} for blind source separation and ICA, in which the encoder is interpreted as a demixing operator and the decoder is interpreted as a remixing or generative mixing operator, as illustrated in Fig.~\ref{fig:structured_prior_vae_bss}. In this framework, each latent dimension is explicitly treated as an independent source component and is assigned its own parameterized prior $p_{\psi}(s_i)$, while the corresponding posterior $q_{\phi}(s_i \mid x)$ is inferred by the encoder and regularized toward the prior through the KL divergence. Such a design allows different latent dimensions to be influenced by different prior constraints, so that they are encouraged to evolve into different source signals rather than collapsing into homogeneous latent representations. More importantly, when the prior parameters $\psi$ associated with different latent components converge to distinct values, this indicates that different structured dependencies---for example, temporal or spatial correlations---are being captured in different source dimensions. This structural-prior viewpoint underlies our previous parameter-adaptive VAE (PAVAE) for BSS, our factorized and structured encoder-free Half-AVAE for underdetermined ICA, and our Structured Kernel Regression VAE as a computationally efficient surrogate for GP-based structured priors (\cite{wei2024innovative,wei2025half,wei2025structured}). 

Recent work has also begun to study autoregressive latent-variable extensions of iVAE for nonlinear and nonstationary spatio-temporal blind source separation, with a particular emphasis on autoregressive latent component models, auxiliary variables, and identifiability analysis \cite{sipila2025identifiable}. In contrast, the present study is positioned primarily as a continuation of our earlier structured-prior VAE line for BSS: rather than introducing autoregressive latent modeling from scratch, we further develop the same structural-prior viewpoint by replacing our earlier GP-based prior with a per-source autoregressive flow prior, thereby enriching the prior family while preserving the interpretation of latent dimensions as source components.

Furthermore, since this line of research is fundamentally built upon a generative modeling architecture, once its BSS capability is established, it also shows clear potential for subsequent developments toward disentanglement, identifiability, and interpretable generative modeling.

\section{Proposed Method: AR-Flow VAE}
\label{sec:method}

\subsection{Overview}
\label{subsec:overview}

We propose \emph{AR-Flow VAE}, a variational autoencoder for blind source separation (BSS) in which each latent source is endowed with a \emph{structured autoregressive flow prior}.
The key idea is to replace the conventional simple latent prior with a per-source prior that explicitly captures \emph{temporally or spatially ordered dependence}, while retaining exact density evaluation through an invertible change of variables.

Let
\[
\bm{X} = \{\bm{x}_r\}_{r=1}^{R}, \qquad \bm{x}_r \in \R^{m},
\]
denote the observed mixtures on an ordered index domain.
Here, the ordered index $r$ may represent time, spatial order, or any serial traversal induced by temporal, spatial, or spatio-temporal structure.
Let
\[
\bm{S} = \{\bm{s}_r\}_{r=1}^{R}, \qquad \bm{s}_r \in \R^{n},
\]
denote the latent source vectors, where $n$ is the number of sources and $m$ is the observation dimension.

Unlike formulations tied to a specific linear architecture, our framework is defined through a general encoder--decoder pair:
\begin{align}
\bm{\mu}_r &= f_{\phi}(\bm{x}_r), \label{eq:general_encoder}\\
\hat{\bm{x}}_r &= g_{\theta}(\bm{s}_r), \label{eq:general_decoder}
\end{align}
where $f_{\phi}(\cdot)$ and $g_{\theta}(\cdot)$ are differentiable mappings parameterized by neural-network parameters $\phi$ and $\theta$, respectively.
In the present implementation, they may be instantiated in a lightweight form for computational efficiency, but the proposed formulation is not restricted to linear mappings and naturally extends to deeper nonlinear architectures.

\subsection{Generative Model}
\label{subsec:generative_model}

The joint model is defined as
\begin{equation}
p_{\theta,\psi}(\bm{X},\bm{S})
=
p_{\theta}(\bm{X}\mid \bm{S})\,p_{\psi}(\bm{S}),
\label{eq:joint_model}
\end{equation}
where $p_{\theta}(\bm{X}\mid \bm{S})$ is the observation model and $p_{\psi}(\bm{S})$ is the structured latent prior.

\paragraph{Observation model.}
Conditioned on the latent sources, the observations are generated independently over the ordered index:
\begin{equation}
p_{\theta}(\bm{X}\mid \bm{S})
=
\prod_{r=1}^{R} p_{\theta}(\bm{x}_r\mid \bm{s}_r).
\label{eq:obs_factorization}
\end{equation}
For continuous observations, we use a Gaussian reconstruction model,
\begin{equation}
p_{\theta}(\bm{x}_r\mid \bm{s}_r)
=
\mathcal{N}\!\bigl(\bm{x}_r;\, g_{\theta}(\bm{s}_r),\, v_y \bm{I}\bigr),
\label{eq:gaussian_decoder}
\end{equation}
where $v_y>0$ is the observation noise variance and $\bm{I}$ is the identity matrix.

Therefore,
\begin{equation}
\log p_{\theta}(\bm{X}\mid \bm{S})
=
-\frac{1}{2v_y}\sum_{r=1}^{R}
\left\|
\bm{x}_r - g_{\theta}(\bm{s}_r)
\right\|_2^2
-\frac{Rm}{2}\log(2\pi v_y).
\label{eq:loglikelihood}
\end{equation}

\subsection{Variational Posterior}
\label{subsec:variational_posterior}

We approximate the intractable posterior $p(\bm{S}\mid\bm{X})$ by a factorized Gaussian variational family:
\begin{equation}
q_{\phi}(\bm{S}\mid \bm{X})
=
\prod_{j=1}^{n}\prod_{r=1}^{R}
\mathcal{N}(s_{r,j};\, \mu_{r,j},\, q_j),
\label{eq:posterior_factorized}
\end{equation}
where
\[
\bm{\mu}_r = f_{\phi}(\bm{x}_r)
\]
is the encoder output and $q_j>0$ is the posterior variance associated with the $j$th source.
The posterior variance is shared across the ordered index $r$ but differs across latent source dimensions.

Using the reparameterization trick,
\begin{equation}
s_{r,j}
=
\mu_{r,j} + \sqrt{q_j}\,\epsilon_{r,j},
\qquad
\epsilon_{r,j}\sim \mathcal{N}(0,1).
\label{eq:reparameterization_scalar}
\end{equation}
In matrix form,
\begin{equation}
\bm{S}
=
\bm{M} + \bm{\Xi}\odot \sqrt{\bm{q}},
\label{eq:reparameterization_matrix}
\end{equation}
where
\[
\bm{M} =
\begin{bmatrix}
\bm{\mu}_1^\top\\
\vdots\\
\bm{\mu}_R^\top
\end{bmatrix}
\in \R^{R\times n},
\qquad
\bm{\Xi}\in \R^{R\times n},
\]
$\bm{\Xi}$ contains i.i.d.\ standard Gaussian entries, $\bm{q}=(q_1,\dots,q_n)$, and $\odot$ denotes element-wise multiplication with broadcasting along the ordered index.

The posterior log-density is
\begin{equation}
\log q_{\phi}(\bm{S}\mid\bm{X})
=
\sum_{j=1}^{n}\sum_{r=1}^{R}
\left[
-\frac{1}{2}\frac{(s_{r,j}-\mu_{r,j})^2}{q_j}
-\frac{1}{2}\log(2\pi)
-\frac{1}{2}\log q_j
\right].
\label{eq:log_variational_posterior}
\end{equation}

\subsection{Per-Source Autoregressive Flow Prior}
\label{subsec:arflow_prior}

To encourage disentangled source recovery, we assume that the latent prior factorizes across source dimensions:
\begin{equation}
p_{\psi}(\bm{S})
=
\prod_{j=1}^{n} p_{\psi}(\bm{s}_{1:R,j}),
\label{eq:per_source_prior}
\end{equation}
where
\[
\bm{s}_{1:R,j} = (s_{1,j}, s_{2,j}, \dots, s_{R,j})^\top
\]
is the ordered trajectory of the $j$th source.

Each latent source is modeled by a first-order autoregressive backbone followed by an invertible autoregressive flow.

\subsubsection{Initial state}
For each source $j$, the initial state is Gaussian:
\begin{equation}
s_{1,j}\sim \mathcal{N}(0,\sigma_{0,j}^2),
\qquad \sigma_{0,j}^2 > 0.
\label{eq:initial_state}
\end{equation}
Thus,
\begin{equation}
\log p(s_{1,j})
=
-\frac{1}{2}
\left(
\frac{s_{1,j}^2}{\sigma_{0,j}^2}
+
\log(2\pi)
+
\log \sigma_{0,j}^2
\right).
\label{eq:log_initial_state}
\end{equation}

\subsubsection{Autoregressive backbone}
For $r=2,\dots,R$, define the normalized autoregressive residual
\begin{equation}
u_{r,j}
=
\frac{s_{r,j} - a_j s_{r-1,j}}{\sigma_j},
\label{eq:ar_backbone}
\end{equation}
where $a_j$ is the AR coefficient and $\sigma_j>0$ is the innovation scale.
Without the flow, $u_{r,j}$ would be modeled as standard Gaussian, yielding a conventional Gaussian AR prior.
Our method instead lets $u_{r,j}$ follow a more expressive distribution generated by a conditional invertible flow.

\subsubsection{Autoregressive flow transform}
For each source $j$, let $\varepsilon_{r,j}$ denote the base noise and let $\bm{h}_{r,j}\in \R^{H}$ denote a hidden state that summarizes previous innovations.
The forward flow is defined as
\begin{equation}
u_{r,j}
=
b_{r,j} + \exp(\alpha_{r,j})\,\varepsilon_{r,j},
\label{eq:flow_forward}
\end{equation}
where
\begin{align}
b_{r,j}
&=
(\bm{w}_{j}^{(b)})^\top \bm{h}_{r-1,j} + c_j^{(b)},
\label{eq:flow_shift}
\\
\alpha_{r,j}
&=
\kappa \tanh\!\Bigl(
(\bm{w}_{j}^{(\alpha)})^\top \bm{h}_{r-1,j} + c_j^{(\alpha)}
\Bigr),
\label{eq:flow_logscale}
\end{align}
with $\kappa\in(0,1)$ controlling the dynamic range of the log-scale.
In our implementation, $\kappa=0.8$.

The hidden state is updated recursively as
\begin{equation}
\bm{h}_{r,j}
=
\tanh\!\Bigl(
\bm{W}_{j}^{(h)}\bm{h}_{r-1,j}
+
\bm{W}_{j}^{(\varepsilon)}\varepsilon_{r,j}
+
\bm{c}_{j}^{(\varepsilon)}
\Bigr),
\label{eq:hidden_state_update}
\end{equation}
with initial condition
\begin{equation}
\bm{h}_{1,j} = \bm{0}.
\label{eq:hidden_init}
\end{equation}

Since Eq.~\eqref{eq:flow_forward} is affine in $\varepsilon_{r,j}$, it is exactly invertible:
\begin{equation}
\varepsilon_{r,j}
=
\bigl(u_{r,j} - b_{r,j}\bigr)\exp(-\alpha_{r,j}).
\label{eq:flow_inverse}
\end{equation}

This construction defines a per-source autoregressive flow prior that can capture non-Gaussian, heavy-tailed, impulsive, and nonlinear ordered dependencies, including temporal and spatially ordered source structures.

\subsection{Exact Prior Density by Change of Variables}
\label{subsec:exact_prior_density}

A key advantage of the proposed prior is that its log-density remains exactly tractable.

\subsubsection{Step 1: Transformation from source trajectory to AR residual}

We first transform the source trajectory segment
\[
\bm{s}_{2:R,j}
\]
into the autoregressive residual vector
\[
\bm{u}_{2:R,j}.
\]

From Eq.~\eqref{eq:ar_backbone},
\[
u_{r,j}
=
\frac{s_{r,j}-a_j s_{r-1,j}}{\sigma_j},
\qquad r=2,\dots,R.
\]
For fixed previous states, the derivative with respect to $s_{r,j}$ is
\[
\frac{\partial u_{r,j}}{\partial s_{r,j}} = \frac{1}{\sigma_j}.
\]
Because the transformation is causal along the ordered index, its Jacobian is triangular:
\begin{equation}
\left|
\det
\frac{\partial \bm{u}_{2:R,j}}{\partial \bm{s}_{2:R,j}}
\right|
=
\sigma_j^{-(R-1)}.
\label{eq:jacobian_ar}
\end{equation}
Hence,
\begin{equation}
\log
\left|
\det
\frac{\partial \bm{u}_{2:R,j}}{\partial \bm{s}_{2:R,j}}
\right|
=
-(R-1)\log \sigma_j.
\label{eq:log_jacobian_ar}
\end{equation}

\subsubsection{Step 2: Transformation from AR residual to flow noise}

We next transform the autoregressive residual vector
\[
\bm{u}_{2:R,j}
\]
into the flow base-noise vector
\[
\bm{\varepsilon}_{2:R,j}.
\]

From Eq.~\eqref{eq:flow_inverse},
\[
\varepsilon_{r,j}
=
(u_{r,j}-b_{r,j})\exp(-\alpha_{r,j}),
\]
thus
\[
\frac{\partial \varepsilon_{r,j}}{\partial u_{r,j}}
=
\exp(-\alpha_{r,j}).
\]
Again, causality makes the full Jacobian triangular:
\begin{equation}
\left|
\det
\frac{\partial \bm{\varepsilon}_{2:R,j}}{\partial \bm{u}_{2:R,j}}
\right|
=
\prod_{r=2}^{R}\exp(-\alpha_{r,j})
=
\exp\!\left(-\sum_{r=2}^{R}\alpha_{r,j}\right).
\label{eq:jacobian_flow}
\end{equation}
Therefore,
\begin{equation}
\log
\left|
\det
\frac{\partial \bm{\varepsilon}_{2:R,j}}{\partial \bm{u}_{2:R,j}}
\right|
=
-\sum_{r=2}^{R}\alpha_{r,j}.
\label{eq:log_jacobian_flow}
\end{equation}

\subsubsection{Step 3: Base density evaluation}

After obtaining the base-noise vector
\[
\bm{\varepsilon}_{2:R,j},
\]
we evaluate its Gaussian log-density under the standard normal base distribution.

We assume the base noise is standard Gaussian:
\begin{equation}
\varepsilon_{r,j}\sim \mathcal{N}(0,1),
\qquad r=2,\dots,R.
\label{eq:base_distribution}
\end{equation}
Thus,
\begin{equation}
\log p(\bm{\varepsilon}_{2:R,j})
=
-\frac{1}{2}\sum_{r=2}^{R}
\left(
\varepsilon_{r,j}^2 + \log(2\pi)
\right).
\label{eq:base_logdensity}
\end{equation}

\subsubsection{Final prior density}
Combining the initial density, base density, and both Jacobian terms, the prior log-density for source $j$ becomes
\begin{equation}
\log p_{\psi}(\bm{s}_{1:R,j})
=
\log p(s_{1,j})
+
\log p(\bm{\varepsilon}_{2:R,j})
+
\log
\left|
\det
\frac{\partial \bm{\varepsilon}_{2:R,j}}{\partial \bm{u}_{2:R,j}}
\right|
+
\log
\left|
\det
\frac{\partial \bm{u}_{2:R,j}}{\partial \bm{s}_{2:R,j}}
\right|.
\label{eq:single_source_logprior_compact}
\end{equation}
Substituting Eqs.~\eqref{eq:log_initial_state}, \eqref{eq:log_jacobian_ar}, \eqref{eq:log_jacobian_flow}, and \eqref{eq:base_logdensity}, we obtain
\begin{align}
\log p_{\psi}(\bm{s}_{1:R,j})
&=
-\frac{1}{2}
\left(
\frac{s_{1,j}^2}{\sigma_{0,j}^2}
+
\log(2\pi)
+
\log \sigma_{0,j}^2
\right)
\nonumber\\
&\quad
-\frac{1}{2}\sum_{r=2}^{R}
\left(
\varepsilon_{r,j}^2 + \log(2\pi)
\right)
-\sum_{r=2}^{R}\alpha_{r,j}
-(R-1)\log \sigma_j.
\label{eq:single_source_logprior_expanded}
\end{align}

Since the prior factorizes across source dimensions,
\begin{equation}
\log p_{\psi}(\bm{S})
=
\sum_{j=1}^{n}
\log p_{\psi}(\bm{s}_{1:R,j}).
\label{eq:full_logprior}
\end{equation}

\subsection{Evidence Lower Bound}
\label{subsec:elbo}

The variational objective is the evidence lower bound (ELBO):
\begin{equation}
\mathcal{L}(\theta,\phi,\psi;\bm{X})
=
\E_{q_{\phi}(\bm{S}\mid \bm{X})}
\Big[
\log p_{\theta}(\bm{X}\mid \bm{S})
+
\log p_{\psi}(\bm{S})
-
\log q_{\phi}(\bm{S}\mid \bm{X})
\Big].
\label{eq:elbo}
\end{equation}

Using one Monte Carlo sample from the reparameterized posterior,
\begin{equation}
\widehat{\mathcal{L}}
=
\log p_{\theta}(\bm{X}\mid \bm{S})
+
\log p_{\psi}(\bm{S})
-
\log q_{\phi}(\bm{S}\mid \bm{X}).
\label{eq:mc_elbo}
\end{equation}

In optimization, we minimize the negative ELBO in the normalized form
\begin{equation}
\mathcal{J}
=
\,\mathrm{Rec}(\bm{X},\hat{\bm{X}})
+
{\beta}
\Big(
\log q_{\phi}(\bm{S}\mid \bm{X})
-
\log p_{\psi}(\bm{S})
\Big),
\label{eq:final_loss}
\end{equation}
where
\begin{equation}
\hat{\bm{x}}_r = g_{\theta}(\bm{s}_r), \qquad r=1,\dots,R,
\label{eq:reconstruction}
\end{equation}
$\mathrm{Rec}(\bm{X},\hat{\bm{X}})$ is the mean-squared reconstruction term, and $\beta>0$ controls the strength of the KL-related regularization.

\subsection{Interpretation}
\label{subsec:interpretation}

The proposed AR-Flow VAE can be interpreted as a structured BSS framework with three coupled components:

\begin{enumerate}
    \item The encoder $f_{\phi}$ infers latent source means from the observed mixtures;
    \item The decoder $g_{\theta}$ reconstructs the observations from sampled latent sources;
    \item The per-source AR-Flow prior imposes temporally or spatially ordered dependence on each latent source while preserving cross-source factorization.
\end{enumerate}

Compared with a standard isotropic Gaussian prior, the proposed prior is substantially more expressive.
Compared with a conventional Gaussian AR prior, it additionally models nonlinear and history-dependent residual structure through an invertible autoregressive flow.
This richer prior can better accommodate latent sources with non-Gaussian, heavy-tailed, impulsive, or otherwise structured ordered behavior.

\subsection{Algorithm}
\label{subsec:algorithm}

Algorithm~\ref{alg:arflowvae} summarizes the overall training procedure of AR-Flow VAE.

\begin{algorithm}[htbp]
\small
\caption{Training Procedure of AR-Flow VAE}
\label{alg:arflowvae}
\DontPrintSemicolon
\SetAlgoLined

\KwIn{Observed mixtures $\bm{X}=\{\bm{x}_r\}_{r=1}^{R}$; number of latent sources $n$}
\KwOut{Trained parameters $(\phi,\theta,\psi)$ and inferred posterior $q_{\phi}(\bm{S}\mid\bm{X})$, parameterized by $\{\bm{\mu}_r\}_{r=1}^{R}$ and $\bm{q}=(q_1,\dots,q_n)$}

\textbf{Parameters:} encoder parameters $\phi$, decoder parameters $\theta$, posterior variance parameters, autoregressive prior parameters, and flow parameters collected in $\psi$\;

Initialize $(\phi,\theta,\psi)$\;

\While{not converged}{
    Compute posterior means $\bm{\mu}_r = f_{\phi}(\bm{x}_r)$ for $r=1,\dots,R$\;
    
    Sample latent sources from the variational posterior
    $s_{r,j} = \mu_{r,j} + \sqrt{q_j}\,\epsilon_{r,j}$, where $\epsilon_{r,j}\sim\mathcal{N}(0,1)$\;
    
    Reconstruct observations $\hat{\bm{x}}_r = g_{\theta}(\bm{s}_r)$ for $r=1,\dots,R$\;
    
    Evaluate the reconstruction term $\log p_{\theta}(\bm{X}\mid\bm{S})$\;
    
    Evaluate the posterior log-density $\log q_{\phi}(\bm{S}\mid\bm{X})$\;
    
    \For{$j \leftarrow 1$ \KwTo $n$}{
        Compute the initial-state log-density $\log p(s_{1,j})$\;
        
        Form autoregressive residuals $u_{r,j} = (s_{r,j}-a_j s_{r-1,j})/\sigma_j$ for $r=2,\dots,R$\;
        
        Apply the inverse autoregressive flow to obtain
        $\varepsilon_{r,j} = (u_{r,j}-b_{r,j})\exp(-\alpha_{r,j})$\;
        
        Compute the per-source prior log-density
        \[
        \log p_{\psi}(\bm{s}_{1:R,j})
        =
        \log p(s_{1,j})
        +
        \log p(\bm{\varepsilon}_{2:R,j})
        -
        \sum_{r=2}^{R}\alpha_{r,j}
        -
        (R-1)\log\sigma_j
        \]
    }
    
    Sum over sources:
    \[
    \log p_{\psi}(\bm{S})
    =
    \sum_{j=1}^{n}\log p_{\psi}(\bm{s}_{1:R,j})
    \]
    
    Form the training objective
    \[
    \mathcal{J}
    =
    \mathrm{Rec}(\bm{X},\hat{\bm{X}})
    +
    \frac{\beta}{mR}
    \bigl(
    \log q_{\phi}(\bm{S}\mid\bm{X})
    -
    \log p_{\psi}(\bm{S})
    \bigr)
    \]
    
    Update $(\phi,\theta,\psi)$ by gradient descent\;
}

\Return $(\phi,\theta,\psi)$ and
$q_{\phi}(\bm{S}\mid\bm{X})=\prod_{r=1}^{R}\prod_{j=1}^{n}\mathcal{N}(s_{r,j};\,\mu_{r,j},\,q_j)$\;
\end{algorithm}

\section{Experimental Study}
\label{sec:simulation}

\subsection{Synthetic Signals}
\label{subsec:synthetic_experiment}

To verify the blind source separation capability of the proposed AR-Flow VAE, we first conduct a simulation study using three synthetic source signals. These source signals are artificially generated to exhibit different ordered structures and are then linearly mixed to form the observed signals. The mixed observations are used as the input to AR-Flow VAE, without providing any source labels or pairing information during training. In this sense, the experiment is fully unsupervised.

\begin{figure}[t]
    \centering
    \includegraphics[width=0.98\linewidth]{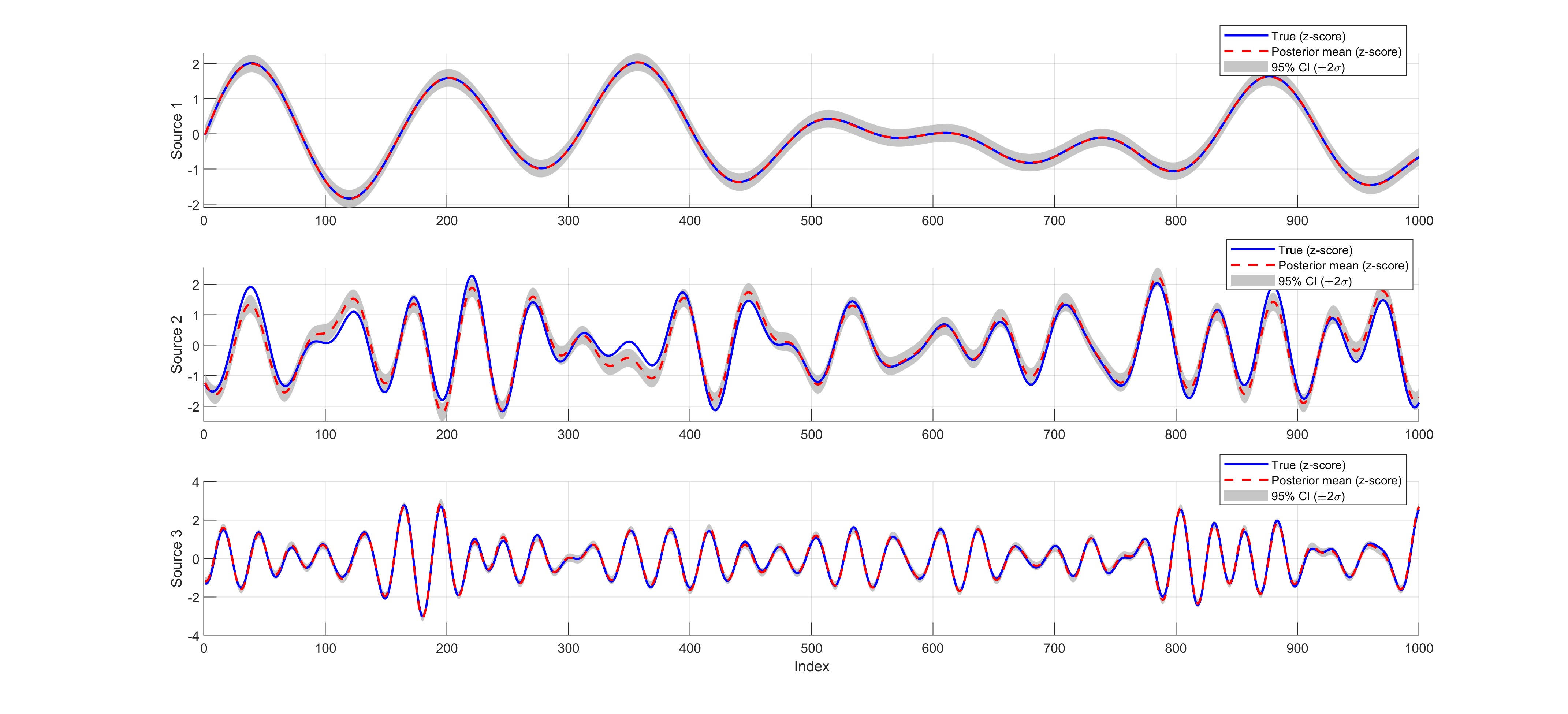}
    \caption{Recovered sources in the synthetic experiment. For each source (Z-score normalized), the true signal, the inferred posterior mean, and the corresponding 95\% confidence interval are shown.}
    \label{fig:synthetic_recovery}
\end{figure}

Figure~\ref{fig:synthetic_recovery} presents the Z-Score normalized source recovery results. It can be seen that the inferred posterior means basically match the true source signals for all three components, indicating that the proposed AR-Flow VAE is able to effectively separate the mixed observations into their underlying sources. In addition, because the proposed framework is formulated as a probabilistic generative model, it provides not only point estimates of the recovered sources but also the corresponding posterior distributions, represented here by the posterior mean and confidence interval. This is an important advantage over purely deterministic separation models, as it allows uncertainty in the inferred sources to be explicitly quantified. Variational autoencoder based probabilistic models are widely used precisely because they approximate posterior distributions rather than producing only deterministic latent estimates. 

\begin{figure}[t]
    \centering
    \includegraphics[width=0.98\linewidth]{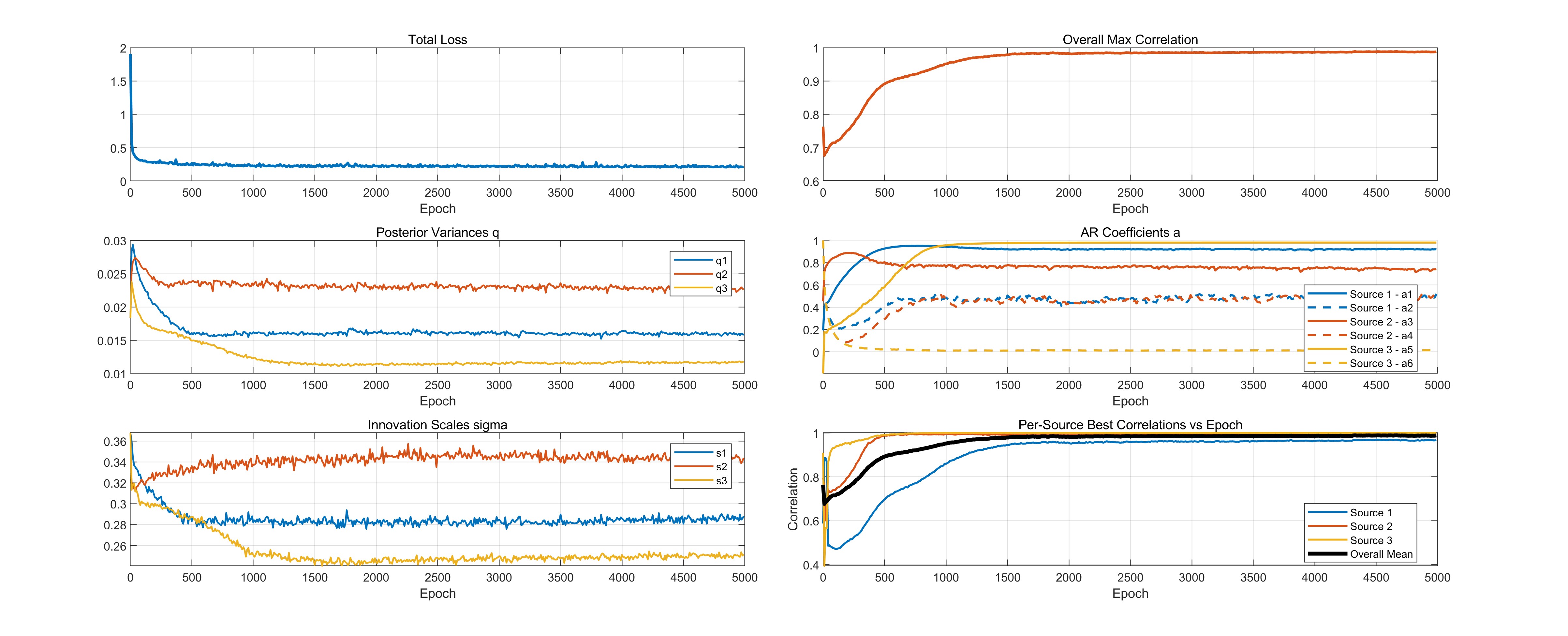}
    \caption{Training traces of AR-Flow VAE in the synthetic experiment, including the total loss, overall maximum correlation, posterior variances, autoregressive parameters, innovation scales, and per-source best correlations.}
    \label{fig:synthetic_training_trace}
\end{figure}

Figure~\ref{fig:synthetic_training_trace} further shows the parameter evolution during training. Since the neural-network parameters inside the flow module are relatively high-dimensional, they are not plotted directly. Instead, the figure focuses on the more interpretable structural parameters, including the posterior variances, autoregressive coefficients, innovation scales, and correlation-based monitoring curves. As training proceeds, the autoregressive parameters associated with different sources gradually converge to different values. This trend is consistent with the evolution of the maximum correlations between the recovered sources and the corresponding true sources. It should be emphasized that these maximum correlations are used only for performance monitoring and visualization; they are not included in the loss function and do not provide supervision to the model.

The results in Fig.~\ref{fig:synthetic_training_trace} suggest that, while optimizing the variational objective alone, the structured AR-Flow priors automatically adapt to different ordered dependency patterns carried by different latent source dimensions. In other words, the prior parameters progressively specialize to model different autocorrelation structures, and this specialization occurs simultaneously with the separation of the latent sources. This behavior provides strong evidence that the proposed framework achieves unsupervised blind source separation through structured probabilistic modeling rather than through external source guidance.

Overall, the synthetic experiment demonstrates two important properties of AR-Flow VAE. First, the model is able to recover multiple latent sources from their mixtures with high fidelity. Second, the structured autoregressive flow priors do not merely serve as regularization terms, but actively drive different latent dimensions toward different structured source components. This supports the central idea of the proposed method: by assigning each latent source an adaptive structured prior, the model can automatically discover and separate different underlying source processes in an unsupervised manner.

\section{Conclusion and Outlook}
\label{sec:conclusion}

In this work, we proposed AR-Flow VAE, a structured variational autoencoder for blind source separation in which each latent source is assigned an adaptive autoregressive flow prior. By interpreting the encoder as a demixing mapping and the decoder as a generative remixing mapping, the proposed framework provides a probabilistic formulation of BSS in which latent source recovery and structured prior learning are carried out jointly. The experimental results on synthetic signals show that AR-Flow VAE can effectively separate mixed observations into their underlying source components, while simultaneously providing posterior distributions for the inferred sources rather than only deterministic point estimates. The parameter evolution further suggests that, during optimization of the variational objective, different latent dimensions are automatically driven toward different structured source processes, which supports the central idea of unsupervised source separation through adaptive structured priors.

Several directions deserve further investigation. First, the computational efficiency of AR-Flow VAE should be improved, especially for longer sequences, larger latent dimensions, and more expressive flow modules. This may involve more efficient flow parameterizations, reduced-complexity sequential updates, partial parallelization strategies, or surrogate structured priors that retain most of the modeling power with lower computational cost. Second, future studies should examine the behavior of AR-Flow VAE on more challenging sources, including strongly nonstationary, heavy-tailed, impulsive, multimodal, spatially correlated, and more highly nonlinear source processes, as well as more difficult mixing conditions. Beyond source separation performance alone, it is also meaningful to further explore whether the proposed architecture can serve as a foundation for subsequent developments in disentanglement, identifiability, and interpretable structured generative modeling.

\clearpage

\bibliography{ref}

\end{document}